%% file: main.tex
\documentclass{Interspeech}

\interspeechcameraready 

\usepackage{amsmath}

\title{``Alexa, can you \textit{forget} me?'' \\Machine Unlearning Benchmark in Spoken Language Understanding} 

\author[affiliation={1}]{Alkis}{Koudounas*}
\author[affiliation={1}]{Claudio}{Savelli*}
\author[affiliation={1}]{Flavio}{Giobergia}
\author[affiliation={1}]{Elena}{Baralis}

\affiliation{}{Politecnico di Torino}{Italy}
\email{name.surname@polito.it}

\usepackage{comment}
\usepackage{cite}
\usepackage{multirow}
\usepackage[table,xcdraw]{xcolor}

\newcommand{\benchmark}{\texttt{UnSLU-BENCH}\xspace}
\newcommand{\numMU}{eight\xspace}
\newcommand{\gum}{$\text{GUM}$\xspace}
\newcommand{\mia}{\text{MIA}\xspace}

\newcommand{\equalcontrib}{$^{*}$}
\newcommand\blfootnote[1]{%
  \begingroup
  \renewcommand\thefootnote{}\footnote{#1}%
  \addtocounter{footnote}{-1}%
  \endgroup
  }

\begin{document}

\maketitle

\begin{abstract}
Machine unlearning, the process of efficiently removing specific information from machine learning models, is a growing area of interest for responsible AI. However, few studies have explored the effectiveness of unlearning methods on complex tasks, particularly speech-related ones.
This paper introduces \benchmark, the first benchmark for machine unlearning in spoken language understanding (SLU), focusing on four datasets spanning four languages. We address the unlearning of data from specific speakers as a way to evaluate the quality of potential \textit{``right to be forgotten''} requests. 
We assess \numMU unlearning techniques and propose a novel metric to simultaneously better capture their efficacy, utility, and efficiency.
\benchmark\ sets a foundation for unlearning in SLU and reveals significant differences in the effectiveness and computational feasibility of various techniques.
\end{abstract}

\section{Introduction}
\blfootnote{\equalcontrib{} Both authors contributed equally to this work.}
Machine unlearning (MU) refers to the process of efficiently removing specific data points from a trained machine learning model without the need for a complete retraining from scratch~\cite{mu_survey}.
This capability is crucial for complying with data privacy regulations, such as the European Union's General Data Protection Regulation (GDPR)~\cite{gdpr} and the California Consumer Privacy Act (CCPA)~\cite{ccpa}, which promote the \textit{``right to be forgotten''}.
By removing the influence of specific data points on machine learning models, MU helps maintain compliance with legal standards and protects user privacy~\cite{xu2024machine}.

In the context of speech, MU plays an even more important role.
Speech data often contains personally identifiable information, making it particularly sensitive~\cite{nautsch2019gdpr, 10626001, koudounas2025privacy}. %removed 9747231
The ability to unlearn specific data ensures that individuals can exercise control over their personal information, thus increasing trust in AI systems.
In addition, unlearning mechanisms can help reduce the influence of unreliable data and mitigate biases, contributing to the development of more fair speech recognition models~\cite{NEURIPS2023_2ecc8008, koudounas2024contrastive, hine2024supporting}.

One important example is the interaction with vocal assistants.
These models process large amounts of user speech data to perform tasks such as intent classification~\cite{ma2024speech}.
Ensuring that these systems can unlearn data from individual users upon request is essential to maintain user autonomy and privacy~\cite{singh2019profiling, mehta2020recent}.  
Despite the critical nature of this capability, there is a non-negligible gap in existing research on MU tailored to speech tasks.
While MU has been explored in other domains, including text~\cite{jang2022knowledge, eldan2023s} and image~\cite{limachine, liu2024machine} processing, its application to speech tasks remains under-developed.

The authors of~\cite{mason-williams2025machine} first explore the application of MU techniques for audio and speech processing. However, their study is limited to audio classification tasks and uses only a single speech dataset focused on keyword spotting, a task semantically much less complex than the intent classification challenges faced in Spoken Language Understanding (SLU). 
This emphasizes the need for MU techniques specifically designed to handle the complexities of SLU tasks.

To fill this gap, we introduce \benchmark, the first comprehensive benchmark for machine unlearning in SLU.
It includes four intent classification datasets in four different languages: Fluent Speech Commands (FSC)~\cite{fsc} and SLURP~\cite{slurp} in English, ITALIC~\cite{italic} in Italian, and SpeechMASSIVE~\cite{speechmassive} in both German and French. 
For each dataset, we evaluate two transformer models, wav2vec 2.0~\cite{w2v2} and HuBERT~\cite{hubert} for English datasets, and XLS-R-128~\cite{xlsr128} and XLS-R-53~\cite{xlsr53} for the other languages.
The latter model has been fine-tuned on Automatic Speech Recognition (ASR) for each target language.

\benchmark offers a complete analysis of the effectiveness of MU techniques across different model architectures and dataset complexities. 
We evaluate \numMU distinct unlearning methods, examining both their effectiveness and computational efficiency in removing specific speakers' data from the models.

Our contributions can be summarized in four points: (1) we introduce the first benchmark for machine unlearning in SLU, with four datasets in four languages and two models per dataset; (2) we evaluate \numMU unlearning techniques, measuring their impact on data removal and model performance; (3) we propose \gum, a novel MU metric considering efficacy, efficiency, and utility of unlearning methods simultaneously; and (4) we provide an in-depth analysis of unlearning performance across datasets, languages, model sizes and architectures.

This benchmark\footnote{\texttt{\href{https://github.com/koudounasalkis/UnSLU-BENCH}{github.com/koudounasalkis/UnSLU-BENCH}}} aims to advance the development of privacy-preserving techniques in speech tasks, facilitating future research on more trustworthy voice assistant systems.

\section{Machine Unlearning}

\subsection{Problem definition}
We assume a given model $\theta$ that has been trained on a SLU dataset $\mathcal{D}$.
Each data point is represented as a triplet $(x, y, s) \in \mathcal{D}$, where $x$ denotes the utterance, $y$ indicates the target intent, and $s$ is the speaker's identity.
We refer to the set of all speakers in the training set as $\mathcal{S}$. 
We now assume that a subset of speakers $\mathcal{S}_f \subset \mathcal{S}$ asks for their data to be deleted.
From a data perspective, this simply implies deleting from the database all samples $\mathcal{D}_f = \{ (x, y, s)\; |\; s \in \mathcal{S}_f) \}$, referred to as the \textit{forget set}. 
However, those samples have affected the learning process of $\theta$. 
We refer to the remaining samples, i.e., $\mathcal{D}_r = \mathcal{D} \setminus \mathcal{D}_f$ as the \textit{retain set}. 
MU is tasked to remove the influence of points in $\mathcal{D}_f$ from $\theta$. 
In other words, MU algorithms produce a new model $\hat{\theta} = \phi(\theta, \mathcal{D}_r, \mathcal{D}_f)$.
As introduced in~\cite{golatkar2020eternal}, we adopt the idea of a \textit{gold model}, i.e., the model $\theta'$ that has been trained using only $\mathcal{D}_r$. The gold model represents MU's ideal target (i.e., we want $\hat{\theta} \approx \theta'$). However, retraining the model from scratch for every \textit{forget} request is generally unfeasible, especially for larger models -- hence the need for unlearning methods.

\subsection{Unlearning methods}
\benchmark\ includes \numMU MU techniques as follows.

\vspace{1mm}
\noindent \textbf{Fine-Tuning (\texttt{FT})} continues to train the model using all $\mathcal{D}_r$ for one epoch. Thus, being $\mathcal{D}_f$ unseen for one additional epoch, it should be less influential than $\mathcal{D}_r$. This method is commonly used as a baseline in the unlearning framework. 

\vspace{1mm}
\noindent \textbf{Negative Gradients (\texttt{NG})} \cite{golatkar2020eternal} finetunes the model using all \(\mathcal{D}_f\) only. Instead of a normal \texttt{FT}, the gradient direction is reversed during the backpropagation to make the model forget \(\mathcal{D}_f\).  

\vspace{1mm}
\noindent \textbf{NegGrad+ (\texttt{NG+})} \cite{choi2023towards,kurmanji2024towards} was proposed as an extension to Negative Gradients to avoid the so-called \textit{``catastrophic forgetting''}, i.e., the destruction of the model's utility. To do this, in addition of \texttt{NG} on \(\mathcal{D}_f\), \texttt{FT} is done on the whole \(\mathcal{D}_r\). 

\vspace{1mm}
\noindent \textbf{Catastrophically forgetting the last $k$ layers (\texttt{CF-$k$})} \cite{goel2022towards} applies \texttt{FT} only to the final $k$ layers of the model. In this way, the unlearning model is faster, as it applies backpropagation on those layers with the most relevant representations and keeps the rest of the network untouched. 

\vspace{1mm}
\noindent \textbf{UNSIR (\texttt{UNSIR})} \cite{tarun2023fast} includes two phases, first it destroys the model (``\textit{impair}''), and then it rebuilds its utility (``\textit{repair}''). In the first, an error-maximizing noise is created for each element of \(\mathcal{D}_f\), which is then used to train the model in combination with \texttt{FT}. The second phase consists of another epoch of \texttt{FT} only.

\vspace{1mm}
\noindent \textbf{Bad Teaching (\texttt{BT})} \cite{chundawat2023can} uses a competent teacher, i.e., a copy of the original model, and an incompetent teacher, i.e., the same model not fine-tuned on the task, in a distillation setup to train a student to behave like the first on \(\mathcal{D}_r\) and like the second on \(\mathcal{D}_f\). We also evaluate a \textbf{light variant (\texttt{BT-L})} of the method with a random prediction generator as the incompetent teacher. 

\vspace{1mm}
\noindent \textbf{SCRUB (\texttt{SCRUB})} \cite{kurmanji2024towards} uses a teacher-student setup with a single teacher, i.e., a copy of the original model. This method combines three different losses: a first loss maximizes student similarity with the teacher on \(\mathcal{D}_r\), a second loss minimizes it on \(\mathcal{D}_f\), while a third task loss improves the final model utility. 

\input{tables/results_fsc_all}

\input{tables/results_slurp_italic}

\subsection{Unlearning metrics}
The evaluation of unlearning algorithms is not trivial. In literature~\cite{hayes2024inexact}, the three main aspects of interest are \textit{efficacy} (whether the unlearning process effectively erased the required information), \textit{efficiency} (how costly the unlearning process is) and \textit{utility} (whether the unlearned model still successfully addresses the original task). 
We argue that all three aspects should be considered at the same time. Ignoring any one of them can lead to trivial solutions.
If we ignore  \textit{efficacy}, the best ``unlearned'' model is simply the original model. Since we are not checking whether the model has actually forgotten anything, this maximizes efficiency (no computational cost) and utility (performance remains the same).
If we ignore \textit{efficiency}, the best solution is to retrain the model from scratch. As we do not consider the cost of retraining, this maximizes efficacy (the model has never seen the forget set) and utility (the model performs as well as possible).
If we ignore \textit{utility}, the best unlearning method is a model that predicts random values. Since we do not care about the quality of the results, this maximizes efficacy (the model does not retain any knowledge of the forget set) and efficiency (no additional computation is needed).

Despite these considerations, very few works in literature account for combinations of some metrics. NoMUS \cite{choi2023towards} considers efficacy and utility together. The work in \cite{llmunlsemeval2025} selects the most effective method given a utility threshold, while \cite{cadet2024deep} chooses the hyperparameter configuration that maximizes efficacy and then evaluates models based on their efficacy. 

In addition, metrics in literature are typically not considered in relation to the \textit{gold model performance}. 
Since MU aims to produce a model that resembles the model retrained from scratch, we argue that it is fundamental to ground all measures to the gold model. 
We acknowledge, of course, that having access to the gold model is a constraint that is generally only met during model validation and not in deployment. 
This is a limitation that affects the entire field of MU, and no general, gold-free solution has been proposed yet. 

In this work, we introduce a new metric, the \textit{Global Unlearning Metric} (\gum), which considers all three aspects simultaneously, with comparisons against the gold model. 
We quantify utility as the similarity in performance between the gold and the unlearned models as $U = 1 - |F1_T^{(g)} - F1_T^{(u)}|$, based on the macro F1 scores\footnote{Other scenarios may require a change in utility function.} on a test set ($F1_T^{(g)}$ and $F1_T^{(u)}$).
We use the \mia (Membership Inference Attack), a commonly adopted metric in unlearning \cite{hayes2024inexact}, to quantify the efficacy of a method.
More specifically, the \mia of the gold model ($\mia^{(g)}$) is the ideal target, whereas the \mia of the original model $\mia^{(o)}$ is the starting point. 
Based on these boundaries, we quantify the efficacy $E$ as:
\begin{equation*}
    E = 1 - \left( \frac{\mia'^{(u)} - \mia'^{(g)}}{ \mia^{(o)} - \mia'^{(g)}} \right)^2\;,
\end{equation*}
\noindent where $\mia'^{(u)} = \min\;\{\mia^{(u)}, \mia^{(o)}\}$ and $\mia'^{(g)} = \min\;\{ \mia^{(g)}, (\mia'^{(u)} + \mia^{(o)})/2 \}$ are saturated versions of the gold and unlearned \mia that guarantee that $E \in [0, 1]$ in edge cases.
The quantity is squared to increase similarities for small gold-unlearned \mia distances. 
Finally, we quantify the efficiency as the ratio of the logarithms of the unlearning time $T^{(u)}$ and gold retraining time $T^{(g)}$.
\begin{equation*}
T = 1 - \frac{log(T^{(u)} + 1)}{log(T^{(g)} + 1)}\; .
\end{equation*}

We define \gum as the weighted harmonic mean between these three quantities:

\begin{equation*}
    GUM = \frac{(1 + \alpha + \beta)U E T}{\alpha ET + \beta UT + UE} \; .
\end{equation*}

\noindent The $\alpha$ and $\beta$ parameters assign different importance to the three quantities. Here, we weigh all quantities equally ($\alpha = \beta = 1$). 

\section{Experimental Setup}

\vspace{1mm}
\noindent \textbf{Datasets.}
\benchmark includes four publicly available datasets: FSC~\cite{fsc} and SLURP~\cite{slurp} for English, ITALIC~\cite{italic} for Italian, and SpeechMASSIVE~\cite{speechmassive} for German and French. The FSC dataset is relatively straightforward, containing 31 intents. In contrast, SLURP, ITALIC, and SpeechMASSIVE are substantially larger, with 60 intents and greater linguistic diversity. ITALIC and SpeechMASSIVE are multilingual extensions of SLURP, covering Italian, and German-French, respectively\footnote{SpeechMASSIVE covers 12 languages, but we focus on German and French only.}. Unlike other datasets, SLURP does not provide speaker-independent splits, which are, however, required by MU techniques to be effective. In fact, the identities present in the retain, forget, and test sets must be exclusive to successfully apply and evaluate unlearning methods.  
To address this, we propose new speaker-independent splits\footnote{These splits are publicly available in our project repository.}. In the following tables, we refer to the new dataset as SLURP*. 
For the other datasets, we use the original splits, with the identities already separated between train and test splits. To create the \textit{forget} set, individuals with at least 100 associated audio samples were randomly taken from each dataset. This ensures that a sufficiently representative number of points were used for training the model for each individual to be forgotten. 
This implies that the size of \(\mathcal{D}_f\) with respect to \(\mathcal{D}_t\) is 2.5--5\% on the different datasets. In this way, we simulate a real case scenario of a possible request to delete one's personal data from a model's training.

\vspace{1mm}
\noindent \textbf{Models.} For each dataset, we fine-tune two transformer models. For the English datasets, we use wav2vec 2.0~\cite{w2v2} and HuBERT~\cite{hubert} in their base sizes. For the multilingual datasets, we use XLS-R 128~\cite{xlsr128} and XLS-R 53~\cite{xlsr53}. The latter is ASR-fine-tuned for the target language (e.g., Italian, German, French).  

\vspace{1mm}
\noindent \textbf{Unlearning Methods.} For each unlearner, we use two different sets of learning rates as parameter tuning depending on how destructive they are. Specifically, we employ 5e-07, 1e-06, and 5e-06 for \texttt{NG}, \texttt{NG+}, \texttt{BT}, \texttt{BT-L}, \texttt{SCRUB}, and 1e-05, 5e-05, and 1e-04 for \texttt{FT}, \texttt{CF-$k$}, \texttt{UNSIR}. For each experiment, we consider the method that achieves the highest utility, efficacy, and efficiency as the best through the use of \gum. 
Moreover, considering that the original implementation of UNSIR was made to forget entire classes within the dataset, we use the version proposed by \cite{choi2023towards}, applicable also to individual samples.

\input{tables/results_speech_massive}

\begin{figure}
    \centering
    \includegraphics[width=\linewidth]{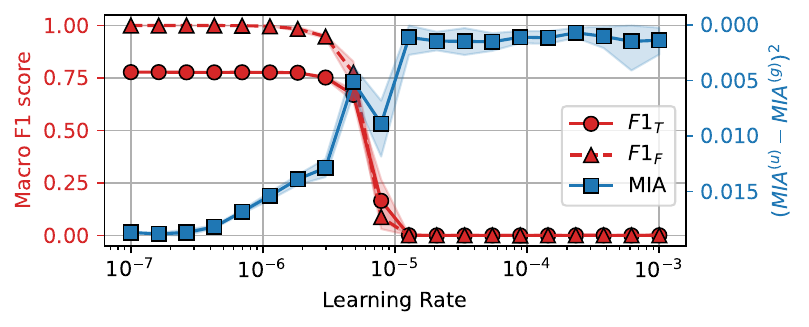}
    \caption{Trade-off between utility (test and forget F1) and efficacy (\mia) on  \texttt{NG}, as the LR changes (ITALIC, XLS-R 53-IT).}
    \label{fig:lr}
    \vspace{-3mm}
\end{figure}

\section{Results}
In the following, we present experiments conducted to explore the behavior of MU techniques in SLU.

\vspace{1mm}
\noindent \textbf{Benchmark results.} The analysis of Tables~\ref{table-fsc}--\ref{table-speech-massive} shows distinct patterns in unlearning methods for MU performance across different models and datasets.
The best F1 and MIA results are measured by their distance from our target's gold model.

\texttt{NG} consistently achieves the highest \gum scores.  
For instance, for wav2vec 2.0, it outperforms the second-best approach by +35\% on FSC and +26\% in SLURP*. For the larger multilingual XLS-R 53 model, it improves \gum by +39\% on ITALIC and SpeechMASSIVE de-DE and by +48\% on SpeechMASSIVE fr-FR. This improvement comes from its exceptional efficiency (speedups up to 1748$\times$ on FSC) and strong efficacy (MIA close to gold models, often ranking first or second among competitors, especially in multilingual datasets).  
\texttt{NG$+$} achieves slightly higher $F1_T$ and $F1_F$ scores than \texttt{NG} in some cases, with comparable MIA scores. However, its overall \gum score is significantly lower as its speedup is one order of magnitude lower than \texttt{NG}.
\texttt{NG$+$}  also suffers the \textit{``catastrophic forgetting''} phenomena in some cases, such as XLS-R 128 ($F1_F$ = .001 on ITALIC, .008 on SpeechMASSIVE fr-FR).  
\texttt{FT} balances utility and efficacy well for complex models.  
For example, XLS-R 128 on ITALIC achieves $F1_T$ = .638, close to the gold model ($F1_T$ = 0.643). However, it is less efficient due to full-network updates, with speedups ranging from 7.96$\times$ to 83.78$\times$.  
CF-$k$ delivers mixed results. It is the second-most efficient method but focuses only on the final layers, which risks incomplete unlearning. This is evident in its higher MIA scores compared to gold models (e.g., .612–.624 vs. gold .493–.520 in SpeechMASSIVE de-DE and fr-FR).
Bad Teaching variants (\texttt{BT}, \texttt{BT-L}) show dataset-dependent performance.  
They achieve good \gum scores on FSC and SLURP but perform poorly on larger multilingual models on ITALIC and SpeechMASSIVE. 
\texttt{SCRUB} and \texttt{UNSIR} perform poorly in \gum, as they achieve moderate speedups (6.21$\times$–65.40$\times$ and 6.55$\times$–64.07$\times$, respectively) but have inconsistent efficacy.  

In conclusion, while most prior works~\cite{grimes2024gone, choi2023towards, jin2024rwku, maini2024tofu} emphasize efficacy and utility, ignoring efficiency, \gum bridges this gap by integrating all three factors. Although more recent alternatives have been proposed, we show that \texttt{NG} remains one of the most well-rounded approaches, performing consistently well across all metrics, as summarized by its large \gum scores.

\vspace{1mm}
\noindent \textbf{(Un)learning rate.} Given a fixed computing budget, the learning rate (LR) is an important parameter influencing the final effect for gradient-based unlearning. 
A small LR implies a lighter effect on the model: the original utility is preserved, but the unlearning effect is limited. Instead, a large LR affects the model more significantly, producing better unlearning, but affecting the overall performance. 
We study this effect empirically for a fixed unlearning technique, \texttt{NG}. 
The trade-off between utility and efficacy is clearly shown in Figure \ref{fig:lr}.

\vspace{1mm}
\noindent \textbf{Advantages of \gum.} 
In Table \ref{tab:gum_nomus}, we compare \gum against NoMUS, the weighted average between model accuracy and \mia \cite{choi2023towards}.
We first note that both Original and Gold models (two trivial ``unlearning'' approaches) achieve large NoMUS scores but obtain -- by definition -- a 0 \gum score.
\texttt{UNSIR} deteriorates the efficacy, with a \mia score worse than the original model. 
As a consequence, the model obtains \gum = 0. 
However, the same method achieves NoMUS = .700. This unexpectedly large value is due to the fact that NoMUS does not contextualize \mia scores w.r.t. gold and original values. 
Finally, \texttt{NG} and \texttt{SCRUB} score similarly in terms of utility ($F1_T$) and efficacy (\mia), resulting in similar NoMUS scores.
However, \texttt{NG} is 1748 times faster than retraining, whereas \texttt{SCRUB} is ``only'' 65 times faster.
This (large!) gap in efficiency is reflected in \gum scores (.563 vs .429). 
\input{tables/ablation_GUM_vs_NOMUS}

\input{tables/ablation_epochs}

\vspace{1mm}
\noindent \textbf{Unlearning in SLURP*.} Table~\ref{tab:ablation_2} finally studies the trade-off between model utility and unlearning efficacy tied to training duration.  We consider SLURP*, and produce various Original models, fine-tuned for different numbers of epochs (5 to 60); then we apply unlearning with \texttt{NG+}. 
At 60 epochs, the unlearned model achieves near-gold utility ($F1_T$ = .696 vs. $F1_T^{(g)}$ .707) but shows limited forgetting: its MIA (.611) is close to the original model one (.628), indicating persistent memorization of the forget set.  
This suggests that the prolonged training creates rigid decision boundaries that retain speaker-specific patterns, making unlearning interventions less effective.
In other words, the model is overfitting the training data, making it harder to forget.
Conversely, shorter training durations (5-15 epochs) show better alignment with the gold model (MIA .480-.538 vs. gold .491-.515).
The ideal operating point appears around 11 epochs -- sufficient training to recover utility ($F1_T$ = .499) while maintaining low memorization risk (\mia = .480), before overfitting dominates.  
This demonstrates that effective MU requires careful calibration of training duration to balance \textit{how well} the model learns with \textit{how permanently} training data gets encoded. 

\section{Conclusion}
This paper introduced \benchmark, a novel benchmark for machine unlearning techniques in SLU. We analyzed eight MU techniques across four datasets and two model architectures and sizes each. 
We also introduced \gum, a new metric that simultaneously evaluates the three key MU targets: efficacy, efficiency, and utility. 
\benchmark provides a foundation for evaluating MU in SLU, highlighting the need for further research to develop more trustworthy voice-based AI systems.

\section{Acknowledgments}
This work is supported by the FAIR - Future Artificial Intelligence Research and received funding from the European Union NextGenerationEU (PIANO NAZIONALE DI RIPRESA E RESILIENZA (PNRR) – MISSIONE 4 COMPONENTE 2, INVESTIMENTO 1.3 – D.D. 1555 11/10/2022, PE00000013) and the spoke ``FutureHPC \& BigData'' of the ICSC - Centro Nazionale di Ricerca in High-Performance Computing, Big Data and Quantum Computing funded by the European Union - NextGenerationEU. 
This manuscript reflects only the authors' views and opinions, neither the European Union nor the European Commission can be considered responsible for them. 

\bibliographystyle{IEEEtran}
\bibliography{mybib}
\end{document}

%% file: tables/results_fsc_all.tex
\begin{table}[!ht]
\addtolength{\tabcolsep}{-0.4em}
\centering
\caption{\textbf{Unlearning on FSC.} $F1_{T}$ denotes macro F1 on test set, while $F1_{F}$ on forget set. Best results (i.e., closest to gold model for F1 and MIA, highest for others) are in \textbf{bold}, second-best \underline{underlined}. \colorbox[HTML]{c4e2e8}{Original} and \colorbox[HTML]{F1E5AC}{gold} models are highlighted.}
\vspace{-2mm}
\label{table-fsc}
\resizebox{\linewidth}{!}{
\begin{tabular}{c|ccccc|ccccc}
\toprule
\multirow{4}{*}{\textbf{Method}} 
    & \multicolumn{10}{c}{\textbf{FSC}} \\
    \cmidrule{2-11}
& \multicolumn{5}{c|}{\textbf{wav2vec 2.0}} 
    & \multicolumn{5}{c}{\textbf{HuBERT}}  \\ 
    \cmidrule{2-11}
& $\mathbf{F1_{T}}$ 
    & \multicolumn{1}{c}{$\mathbf{F1_{F}}$} 
    & \textbf{MIA} 
    & \multicolumn{1}{c}{\textbf{GUM}} 
    & \textbf{Speedup} 
    & $\mathbf{F1_{T}}$ 
    & $\mathbf{F1_{F}}$ 
    & \textbf{MIA} 
    & \textbf{GUM} 
    & \textbf{Speedup} \\ 
    \midrule
\rowcolor[HTML]{c4e2e8}\texttt{Orig.}
    & .994 
    & 1.00 
    & .508 
    & .000 
    & 1.00$\times$ 
    & .993
    & 1.00
    & .511
    & .000
    & 1.00$\times$ \\
    
\rowcolor[HTML]{F1E5AC}\texttt{Gold}
    & .993 
    & .997 
    & .503 
    & .000 
    & 1.00$\times$ 
    & .991
    & .996
    & .507
    & .000
    & 1.00$\times$ \\
    \midrule
\texttt{FT}
    & \textbf{.993}
    & \underline{.999} 
    & \textbf{.504} 
    & .517 
    & 7.960$\times$ 
    & .979
    & .993
    & \textbf{.508}
    & .514
    & 7.690$\times$ \\    
\texttt{NG}
    & .987 
    & .976 
    & \underline{.501} 
    & \textbf{.816} 
    & \textbf{206.9}$\times$ 
    & \textbf{.992} 
    & \textbf{.996}
    & .514
    & .000
    & \textbf{201.1}$\times$ \\
\texttt{NG$+$}
    & \underline{.994} 
    & .994 
    & .493 
    & .000 
    & 4.030$\times$ 
    & .979
    & .929
    & .510
    & .336
    & 3.900$\times$ \\
\texttt{CF-$k$}
    & \underline{.994} 
    & 1.00 
    & \underline{.501} 
    & \underline{.606} 
    & \underline{16.97}$\times$ 
    & \underline{.993}
    & 1.00
    & \underline{.505}
    & \textbf{.642}
    & \underline{26.70}$\times$ \\
\texttt{UNSIR}
    & .991
    & 1.00 
    & .506 
    & .447 
    & 6.550$\times$ 
    & \underline{.994}
    & .998
    & \textbf{.508}
    & \underline{.484}
    & 6.380$\times$ \\
\texttt{BT}
    & \textbf{.993} 
    & 1.00 
    & .508
    & .000 
    & 4.780$\times$ 
    & \underline{.993}
    & .999
    & .504
    & .363
    & 4.650$\times$ \\
\texttt{BT-L}
    & \underline{.994} 
    & \textbf{.996} 
    & .506 
    & .431 
    & 5.870$\times$ 
    & \underline{.993}
    & \underline{.997}
    & \textbf{.506}
    & .464
    & 5.690$\times$ \\
\texttt{SCRUB}
    & \underline{.994} 
    & 1.00 
    & .506 
    & .439 
    & 6.210$\times$ 
    & \underline{.993}
    & .998
    & \textbf{.508}
    & .479
    & 6.220$\times$ 
    \\
    \bottomrule
\end{tabular}}
\vspace{-3mm}
\end{table}

%% file: tables/results_slurp_italic.tex
\begin{table*}[!ht]
\addtolength{\tabcolsep}{-0.4em}
\centering
\caption{\textbf{Comparison of unlearning methods on SLURP* and ITALIC.} Best results are in \textbf{bold}, second-best \underline{underlined}. 
%\colorbox[HTML]{c4e2e8}{Original} and \colorbox[HTML]{F1E5AC}{gold} models are highlighted.
}
\vspace{-2mm}
\label{table-slurp-italic}
\resizebox{\linewidth}{!}{
\begin{tabular}{c|ccccc|ccccc|ccccc|ccccc}
\toprule
\multirow{4}{*}{\textbf{Method}} 
    & \multicolumn{10}{c|}{\textbf{SLURP*}}
    & \multicolumn{10}{c}{\textbf{ITALIC}} \\
    \cmidrule{2-21}
& \multicolumn{5}{c|}{\textbf{wav2vec 2.0}} 
    & \multicolumn{5}{c|}{\textbf{HuBERT}} 
    & \multicolumn{5}{c|}{\textbf{XLS-R 128}} 
    & \multicolumn{5}{c}{\textbf{XLS-R 53-IT}} \\ 
    \cmidrule{2-21}

& $\mathbf{F1_{T}}$ 
    & \multicolumn{1}{c}{$\mathbf{F1_{F}}$} 
    & \textbf{MIA} 
    & \multicolumn{1}{c}{\textbf{GUM}} 
    & \textbf{Speedup} 
    & $\mathbf{F1_{T}}$ 
    & $\mathbf{F1_{F}}$ 
    & \textbf{MIA} 
    & \textbf{GUM} 
    & \textbf{Speedup} 
    & $\mathbf{F1_{T}}$ 
    & $\mathbf{F1_{F}}$ 
    & \textbf{MIA} 
    & \textbf{GUM} 
    & \textbf{Speedup} 
    & $\mathbf{F1_{T}}$ 
    & $\mathbf{F1_{F}}$ 
    & \textbf{MIA} 
    & \textbf{GUM} 
    & \textbf{Speedup} \\ 
    \midrule

\rowcolor[HTML]{c4e2e8} 
\texttt{Orig.} 
    & .689 
    & 1.000 
    & .628 
    & .000 
    & 1.000$\times$ 
    & .712 
    & 1.000 
    & .613 
    & .000 
    & 1.000$\times$ 
    & .688 
    & .894 
    & .632 
    & .000 
    & 1.000$\times$ 
    & .778 
    & 1.000 
    & .615 
    & .000 
    & 1.000$\times$ \\

\rowcolor[HTML]{F1E5AC} 
\texttt{Gold} 
    & .707 
    & .711 
    & .506 
    & .000 
    & 1.000$\times$ 
    & .704 
    & .715 
    & .492 
    & .000 
    & 1.000$\times$
    & .643 
    & .568 
    & .532 
    & .000 
    & 1.000$\times$ 
    & .784 
    & .736 
    & .478 
    & .000 
    & 1.000$\times$ \\
    \midrule

\texttt{FT}
    & .638 
    & \textbf{.970} 
    & .648 
    & .000 
    & 83.78$\times$ 
    & .734 
    & 1.000 
    & .611 
    & .088 
    & 79.00$\times$ 
    & \textbf{.638} 
    & .671 
    & .555 
    & \underline{.590} 
    & 30.80$\times$ 
    & .711 
    & .850 
    & \underline{.550}
    & \underline{.551}
    & 31.10$\times$ \\

\texttt{NG}
    & .695 
    & \underline{.986} 
    & \underline{.604} 
    & \textbf{.563} 
    & \textbf{1748}$\times$ 
    & .718 
    & .959 
    & .587 
    & \textbf{.587} 
    & \textbf{1654}$\times$ 
    & .679 
    & .868 
    & .603 
    & \textbf{.646} 
    & \textbf{613.4}$\times$ 
    & .590 
    & \underline{.621}
    & \textbf{.525} 
    & \textbf{.766} 
    & \textbf{623.0}$\times$ \\

\texttt{NG$+$}
    & .701 
    & .995 
    & \textbf{.603} 
    & \underline{.446} 
    & 41.63$\times$ 
    & .630 
    & \textbf{.852} 
    & \textbf{.453} 
    & \underline{.578} 
    & 39.30$\times$ 
    & .658 
    & .001 
    & .932 
    & .000 
    & 15.14$\times$ 
    & .743 
    & .936 
    & .582 
    & .418 
    & 15.37$\times$ \\
\texttt{CF-$k$}
    & \textbf{.709} 
    & 1.000 
    & .626 
    & .089 
    & \underline{291.9}$\times$ 
    & .715 
    & 1.000 
    & .608 
    & .196 
    & \underline{274.2}$\times$ 
    & .677 
    & .871 
    & .626 
    & .253 
    & \underline{98.59}$\times$ 
    & \textbf{.781}
    & 1.000 
    & .609 
    & .201 
    & \underline{98.99}$\times$ \\
\texttt{UNSIR}
    & .673 
    & 1.000 
    & .637 
    & .000 
    & 64.07$\times$    
    & .722 
    & 1.000 
    & .613 
    & .000 
    & 60.44$\times$
    & \underline{.636} 
    & .830 
    & .621 
    & .328 
    & 22.01$\times$ 
    & \underline{.775} 
    & 1.000 
    & .612 
    & .109 
    & 22.26$\times$ \\   
\texttt{BT}
    & \underline{.710} 
    & .999 
    & .619 
    & .275 
    & 50.35$\times$ 
    & \underline{.711} 
    & 1.000 
    & .613 
    & .000 
    & 47.42$\times$ 
    & .683 
    & \textbf{.639} 
    & .481 
    & .504 
    & 17.90$\times$ 
    & .731 
    & \textbf{.848} 
    & .557 
    & .491 
    & 17.94$\times$ \\

\texttt{BT-L}
    & .680 
    & .995
    & .637 
    & .000
    & 61.74$\times$ 
    & .685 
    & \underline{.907}
    & \underline{.558} 
    & \underline{.578}
    & 58.11$\times$ 
    & .686 
    & \underline{.651}
    & \underline{.518} 
    & .558
    & 22.02$\times$ 
    & .729 
    & .876
    & .564 
    & .499
    & 22.21$\times$ \\

\texttt{SCRUB}
    & .697 
    & .999 
    & .608 
    & .429 
    & 64.82$\times$ 
    & \textbf{.704} 
    & 1.000 
    & .600 
    & .350 
    & 65.40$\times$ 
    & .442 
    & .357 
    & \textbf{.533} 
    & .536 
    & 23.25$\times$ 
    & .770 
    & .990 
    & .610 
    & .164 
    & 22.66$\times$ \\
    
\bottomrule
\end{tabular}}
\vspace{-3mm}
\end{table*}

%% file: tables/results_speech_massive.tex
\begin{table*}[!ht]
\addtolength{\tabcolsep}{-0.4em}
\centering
\caption{\textbf{Comparison of unlearning methods on SpeechMASSIVE de-DE and fr-FR.} Best results are in \textbf{bold}, second-best \underline{underlined}. 
%\colorbox[HTML]{c4e2e8}{Original} and \colorbox[HTML]{F1E5AC}{gold} models are highlighted.
}
\vspace{-2mm}
\label{table-speech-massive}
\resizebox{\linewidth}{!}{
\begin{tabular}{c|ccccc|ccccc|ccccc|ccccc}
\toprule
\multirow{4}{*}{\textbf{Method}} 
    & \multicolumn{10}{c|}{\textbf{de-De}}
    & \multicolumn{10}{c}{\textbf{fr-FR}} \\
    \cmidrule{2-21}
& \multicolumn{5}{c|}{\textbf{XLS-R 128}} 
    & \multicolumn{5}{c|}{\textbf{XLS-R 53-DE}} 
    & \multicolumn{5}{c|}{\textbf{XLS-R 128}} 
    & \multicolumn{5}{c}{\textbf{XLS-R 53-FR}} \\ 
    \cmidrule{2-21}

& $\mathbf{F1_{T}}$ 
    & \multicolumn{1}{c}{$\mathbf{F1_{F}}$} 
    & \textbf{MIA} 
    & \multicolumn{1}{c}{\textbf{GUM}} 
    & \textbf{Speedup} 
    & $\mathbf{F1_{T}}$ 
    & $\mathbf{F1_{F}}$ 
    & \textbf{MIA} 
    & \textbf{GUM} 
    & \textbf{Speedup} 
    & $\mathbf{F1_{T}}$ 
    & $\mathbf{F1_{F}}$ 
    & \textbf{MIA} 
    & \textbf{GUM} 
    & \textbf{Speedup} 
    & $\mathbf{F1_{T}}$ 
    & $\mathbf{F1_{F}}$ 
    & \textbf{MIA} 
    & \textbf{GUM} 
    & \textbf{Speedup} \\ 
    \midrule

\rowcolor[HTML]{c4e2e8} 
\texttt{Orig.} 
    & .584 
    & .841 
    & .621 
    & .000 
    & 1.000$\times$ 
    & .778 
    & 1.000 
    & .622 
    & .000 
    & 1.000$\times$ 
    & .410 
    & .572 
    & .629 
    & .000 
    & 1.000$\times$ 
    & .756 
    & 1.000 
    & .635 
    & .000 
    & 1.000$\times$ \\

\rowcolor[HTML]{F1E5AC} 
\texttt{Gold} 
    & .566 
    & .529 
    & .513 
    & .000 
    & 1.000$\times$ 
    & .745 
    & .706 
    & .493 
    & .000 
    & 1.000$\times$ 
    & .469 
    & .460 
    & .509 
    & .000 
    & 1.000$\times$ 
    & .772 
    & .800 
    & .520 
    & .000 
    & 1.000$\times$ \\
    \midrule
\texttt{FT}
    & .498 
    & \textbf{.548} 
    & \underline{.543} 
    & \underline{.588} 
    & 34.34$\times$ 
    & .661 
    & \underline{.905} 
    & \underline{.585}
    & \underline{.464} 
    & 17.79$\times$ 
    & .400 
    & \textbf{.465} 
    & \textbf{.539} 
    & \underline{.545}
    & 18.12$\times$ 
    & .759 
    & .974 
    & .627 
    & .255 
    & 18.42$\times$ \\

\texttt{NG}
    & \underline{.550} 
    & .726 
    & .562 
    & \textbf{.797} 
    & \textbf{1078$\times$} 
    & .764 
    & .957 
    & .587 
    & \textbf{.643} 
    & \textbf{558.7$\times$} 
    & .317 
    & .349 
    & \underline{.564} 
    & \textbf{.749} 
    & \textbf{597.3$\times$} 
    & .768 
    & \textbf{.935} 
    & \textbf{.617} 
    & \textbf{.501} 
    & \textbf{610.2$\times$} \\

\texttt{NG$+$}
    & .540 
    & \underline{.567} 
    & \textbf{.487} 
    & .522 
    & 16.89$\times$ 
    & \textbf{.759} 
    & \textbf{.878} 
    & \textbf{.568} 
    & .431 
    & 8.770$\times$ 
    & .382 
    & .008 
    & .882 
    & .000 
    & 8.900$\times$ 
    & .759 
    & \underline{.943} 
    & \underline{.620} 
    & .317 
    & 9.230$\times$ \\    
\texttt{CF-$k$}
    & .587 
    & .865 
    & .622 
    & .000 
    & \underline{109.9$\times$} 
    & .777 
    & 1.000 
    & .616 
    & .208 
    & \underline{56.93$\times$} 
    & \textbf{.436} 
    & .594 
    & .612 
    & .414 
    & \underline{58.23$\times$} 
    & \underline{.770} 
    & 1.000 
    & .624 
    & \underline{.338} 
    & \underline{58.86$\times$} \\   
\texttt{UNSIR}
    & \textbf{.565} 
    & .788 
    & .616 
    & .197 
    & 27.46$\times$   
    & .785 
    & 1.000 
    & .619 
    & .114 
    & 14.23$\times$
    & \underline{.420} 
    & .591 
    & .620 
    & .259 
    & 14.67$\times$ 
    & .768 
    & 1.000 
    & .633 
    & .089 
    & 14.94$\times$ \\    
\texttt{BT}
    & .584 
    & .789 
    & .582 
    & .489 
    & 20.02$\times$ 
    & .726 
    & .945 
    & \underline{.585} 
    & .418 
    & 10.41$\times$ 
    & .411 
    & .583 
    & .597 
    & .409 
    & 10.60$\times$ 
    & \textbf{.772} 
    & .981 
    & .621 
    & .317 
    & 10.82$\times$ \\

\texttt{BT-L}
    & .584 
    & .786
    & .576 
    & .523
    & 24.87$\times$ 
    & \underline{.729} 
    & .948
    & .587 
    & .434
    & 12.94$\times$ 
    & .412 
    & .574
    & .591 
    & .447
    & 13.18$\times$ 
    & .727 
    & .981
    & .623 
    & .306
    & 13.42$\times$ \\

\texttt{SCRUB}
    & .584 
    & .780 
    & .600 
    & .429 
    & 26.86$\times$ 
    & .781 
    & 1.000 
    & .615 
    & .211 
    & 13.43$\times$ 
    & .409 
    & \underline{.532} 
    & .611 
    & .358 
    & 13.68$\times$ 
    & .769 
    & 1.000 
    & .633 
    & .089 
    & 13.94$\times$ \\
    \bottomrule
\end{tabular}}
\vspace{-3mm}
\end{table*}

%% file: tables/ablation_GUM_vs_NOMUS.tex
\begin{table}[]
\centering
\caption{\textbf{Unlearning metrics on SLURP*, wav2vec 2.0.}}
\label{tab:gum_nomus}
\resizebox{.8\linewidth}{!}{
\begin{tabular}{c|ccc|cc}
\toprule
\textbf{Method} & $\mathbf{F1_{T}}$  & \textbf{MIA}  & \textbf{Speedup}                    & \textbf{NoMUS} & \textbf{GUM}  \\
\midrule
\rowcolor[HTML]{c4e2e8} \texttt{Orig.}   & .689 & .628 & 1.000$\times$ & .717  & .000 \\
\rowcolor[HTML]{F1E5AC} \texttt{Gold}    & .707 & .506 & 1.000$\times$ & .848  & .000 \\
\midrule
\texttt{NG}      & .695 & .604 & 1748$\times$  & .744  & .563 \\
\texttt{UNSIR}     & .673 & .637 & 64.07$\times$ & .700  & .000 \\
\texttt{SCRUB}      & .697 & .608 & 64.82$\times$ & .741 & .429 \\
\bottomrule
\end{tabular}}
\vspace{-1mm}
\end{table}

%% file: tables/ablation_epochs.tex
\begin{table}[]
    \centering
    \caption{\textbf{Variation in the difficulty of unlearning as the number of training epochs changes, wav2vec 2.0, SLURP*}. Each experiment uses \texttt{NG+} with LR = 5e-07.}
    \label{tab:ablation_2}
    \resizebox{.85\linewidth}{!}{
    \begin{tabular}{ccccccc}
        \toprule
        \textbf{Epochs} 
            & $\mathbf{F1_T}$ 
            & \cellcolor[HTML]{F1E5AC}$\mathbf{F1_T^{(g)}}$ 
            & \textbf{\mia }
            % & \cellcolor[HTML]{F1E5AC}$\mathbf{\mia^{(g)}}$ 
            & \cellcolor[HTML]{F1E5AC}\textbf{\mia \textsuperscript{(g)}}
            & \cellcolor[HTML]{c4e2e8}\textbf{\mia \textsuperscript{(o)}}
            & \textbf{\gum} \\
        \midrule
        5 
            & .395 
            & \cellcolor[HTML]{F1E5AC}.398 
            & .496  
            & \cellcolor[HTML]{F1E5AC}.510 
            & \cellcolor[HTML]{c4e2e8}.561 
            & .678 \\
        7 
            & .383 
            & \cellcolor[HTML]{F1E5AC}.419 
            & .524  
            & \cellcolor[HTML]{F1E5AC}.515 
            & \cellcolor[HTML]{c4e2e8}.566 
            & .680 \\
        11 
            & .499 
            & \cellcolor[HTML]{F1E5AC}.487 
            & .480 
            & \cellcolor[HTML]{F1E5AC}.492 
            & \cellcolor[HTML]{c4e2e8}.593 
            & .686 \\
        15 
            & .564 
            & \cellcolor[HTML]{F1E5AC}.550 
            & .538 
            & \cellcolor[HTML]{F1E5AC}.491 
            & \cellcolor[HTML]{c4e2e8}.589 
            & .644 \\
        60 
            & .696 
            & \cellcolor[HTML]{F1E5AC}.707 
            & .611 
            & \cellcolor[HTML]{F1E5AC}.506 
            & \cellcolor[HTML]{c4e2e8}.628 
            & .421 \\
        \bottomrule
    \end{tabular}}
    \vspace{-3mm}
\end{table}